\documentclass[a4paper,fleqn,review]{cas-sc}

\usepackage[authoryear]{natbib}

\def\tsc#1{\csdef{#1}{\textsc{\lowercase{#1}}\xspace}}
\tsc{WGM}
\tsc{QE}

\usepackage{subcaption}

\usepackage{xcolor}
\usepackage{listings}
\DeclareCaptionFont{white}{\color{white}}
\DeclareCaptionFormat{listing}{%
  \parbox{\textwidth}{\colorbox{gray}{\parbox{\textwidth}{#3}}\vskip-4pt}}
\makeatletter
\newcommand\notsotiny{\@setfontsize\notsotiny\@vipt\@viipt}
\makeatother

\begin{document}
\let\WriteBookmarks\relax
\def\floatpagepagefraction{1}
\def\textpagefraction{.001}

\shorttitle{RPA-Check: A Multi-Stage Automated Framework for Evaluating Dynamic LLM-based Role-Playing Agents}    

\shortauthors{R. Rosati et~al.}  

\title[mode = title]{RPA-Check: A Multi-Stage Automated Framework for Evaluating Dynamic LLM-based Role-Playing Agents}



\author[1]{Riccardo Rosati}[orcid=0000-0003-3288-638X]
\cormark[1]
\ead{riccardo.rosati@unimc.it}

\credit{Conceptualization, Methodology, Writing – original draft, Supervision}

\affiliation[1]{organization={Department of Political Sciences, Communication and International Relations, University of Macerata},
                addressline={Via Don Minzoni 22/A}, 
                city={Macerata},
                postcode={62100}, 
                country={Italy}}

\author[2]{Edoardo Colucci}
\ead{S1110494@studenti.univpm.it}

\credit{Software, Writing – review and editing}

\affiliation[2]{organization={Deparment of Information Engineering, Università Politecnica delle Marche},
                addressline={Via Brecce Bianche 12}, 
                city={Ancona},
                postcode={60131}, 
                country={Italy}}

\author[2]{Massimiliano Bolognini}
\ead{S1111836@studenti.univpm.it}

\credit{Software, Writing – review and editing}

\author[2]{Adriano Mancini}[orcid=0000-0001-5281-9200]
\ead{a.mancini@staff.univpm.it}

\credit{Writing – review and editing, Supervision}

\author[3]{Paolo Sernani}[orcid=0000-0001-7614-7154]
\ead{paolo.sernani@unimc.it}

\credit{Conceptualization, Methodology, Writing – original draft}

\affiliation[3]{organization={Department of Law, University of Macerata},
                addressline={Piaggia dell'Università 2}, 
                city={Macerata},
                postcode={62100}, 
                country={Italy}}

\cortext[1]{Corresponding author}



\begin{abstract}
The rapid adoption of Large Language Models (LLMs) in interactive systems has enabled the creation of dynamic, open-ended Role-Playing Agents (RPAs). However, evaluating these agents remains a significant challenge, as standard NLP metrics fail to capture the nuances of role adherence, logical consistency, and long-term narrative stability. This paper introduces RPA-Check, a multi-stage automated evaluation framework designed to objectively assess the performance of LLM-based RPAs in complex, constraints-heavy environments. Our methodology is based on a four-step pipeline: (1) Dimension Definition, establishing high-level qualitative behavioral criteria; (2) Augmentation, where these requirements are expanded into granular boolean checklist indicators; (3) Semantic Filtering, to ensure indicator objectivity, no redundancy and agent isolation; and (4) LLM-as-a-Judge Evaluation, which employs chain-of-thought verification to score agent fidelity. 
We validate this framework by applying it to \textit{LLM Court}, a serious game for forensic training involving several quantized local models. Experimental results across five distinct legal scenarios demonstrate the framework’s ability to identify subtle trade-offs between model size, reasoning depth, and operational stability. Notably, the findings reveal an inverse relationship between parametric scale and procedural consistency, showing that smaller, adequately instruction-tuned models (8-9B) can outperform larger architectures prone to user-alignment bias or sycophancy. RPA-Check thus provides a standardized and reproducible metric for future research in generative agent evaluation within specialized domains.
\end{abstract}



\begin{keywords}
Role-Playing Agents \sep Large Language Models \sep Automated Evaluation \sep Serious Games \sep Generative AI
\end{keywords}

\maketitle

\section{Introduction}\label{sec:intro}

The increasing adoption of Large Language Models (LLMs) in interactive systems has enabled the creation of dynamic and open Role-Playing Agents (RPAs) capable of sustaining natural language dialogues without predefined scripts~\citep{Chen2024,Wang2026}. However, evaluating these agents remains a significant challenge: standard Natural Language Processing metrics, such as BLEU, ROUGE or perplexity, do not capture critical dimensions such as adherence to the assigned role, internal logical consistency, and narrative stability over extended conversations. This methodological gap is particularly evident in highly specialized contexts, where correctness does not coincide with linguistic fluency and general factuality, but depends on compliance with procedural and semantic constraints specific to the domain, and players might experience unnatural conversational flow~\citep{Zargham2026}.

In serious games, where one of the goals is to convey professional skills through interactive simulations, the integration of LLM opens opportunities for adaptive and immersive training environments~\citep{Bonetti2025,Ozkaya2025}. However, the lack of objective and reproducible evaluation frameworks limits the ability to compare models and identify trade-offs between different performance dimensions. Existing methodologies, such as the MMLU benchmark~\citep{Hendrycks2021} or Chatbot Arena~\citep{Chiang2024}, focus on general knowledge or conversational utility, making them inadequate for agents who must embody rigidly defined roles, such as an impartial judge or a hostile witness in a mock trial.

This paper introduces RPA-Check, a multi-stage automated evaluation framework designed to systematically and objectively measure the performance of LLM-based RPAs in constrained environments with high procedural complexity. The proposed methodology adapts the CheckEval paradigm~\citep{Pereira2024} to role-playing contexts through a four-stage pipeline: i) Dimension Definition establishes high-level qualitative criteria (such as behavioral role fidelity); ii) Augmentation, in which high-level requirements are expanded into granular Boolean indicators; iii) Filtering, which ensures the objectivity, non-redundancy, and isolation of agent-specific performance metrics; and iv) LLM-as-a-Judge~\citep{Gu2026} Evaluation, where an evaluator model applies chain-of-thought~\citep{Wei2022} checks to quantify the performance of game agents.

The proposed RPA-Check framework is tested on “LLM Court” (also introduced in this paper), a game simulating a court using local quantized models as the LLMs implementing the Non-Playing Characters (NPCs) and developed in Unity. The architecture allows for the execution of trial simulations in which LLM-controlled agents embody legal roles (judge, prosecutor, witnesses) and interact with the player through open, unscripted dialogues. The simulation environment imposes strict procedural constraints, requiring agents to maintain narrative consistency, respect intervention hierarchies, and provide logically consistent responses over extended dialogue sessions. Moreover, the adoption of quantized models allows for completely local execution on consumer hardware~\citep{Lang2024}, eliminating dependence on cloud services, ensuring the privacy of sensitive data, and significantly reducing the computational costs and environmental impact associated with inference~\citep{Bai2024}. This architectural choice enables sustainable and accessible deployment, which is particularly relevant for educational applications intended for contexts with limited resources or confidentiality requirements. 

The experimental results, conducted on seven open-weight models ranging in size from 8 to 14 billion parameters and evaluated across five distinct legal scenarios, highlight the framework's ability to quantify trade-offs between role fidelity, operational stability, and reasoning depth. As such, the proposed framework can quantify abstract qualities like ``Role Adherence'' and ``Logical Consistency,'' distinguishing between models that merely generate fluent text and those that truly embody their assigned roles. In particular, in LLM court, it emerges that smaller but adequately instruction-tuned models can outperform larger architectures in terms of behavioral consistency when measured using a checklist of granular indicators. These results provide useful empirical evidence to guide model selection in similar application contexts, while highlighting limitations related to the sensitivity of models to the specific case and generation seed.

Specifically, this work's contributions to the state of the art are:
\begin{itemize}
    \item RPA-Check, a rigorous and adaptable methodological framework that transforms qualitative role descriptions into quantitative Boolean metrics through an automated pipeline of augmentation, filtering, and evaluation using LLM-as-a-Judge.
    \item LLM Court, a game designed to validate the framework in a rule-based environment that simulates forensic interactions in court under strict procedural constraints. The game introduces a clear separation between the generative logic to produce contents and procedural control, i.e., the courtroom rules (unlike the open-ended conversational agents used in purely recreational contexts).
    \item A modular reference architecture, the one developed for LLM court, that integrates local quantized models into a Unity environment for running process simulations, providing a reproducible testbed for evaluating RPA in constrained contexts.
    \item An empirical comparative analysis of open-weight models that quantifies the trade-off between model size, role fidelity, and operational stability, demonstrating that instruction tuning and evaluation granularity are critical factors for selecting appropriate architectures.
\end{itemize}

The remainder of the paper is organized as follows. Section~\ref{sec:rel} examines the state of the art in the use of LLMs in recreational and educational contexts, methodologies to evaluate dialogic agents, and serious games based on generative artificial intelligence, highlighting the contributions of the proposed work in a comparative manner. Section~\ref{sec:sysDes} describes the system architecture of LLM Court, detailing the technological components and design choices. Section~\ref{sec:method} illustrates the proposed multi-stage automated evaluation framework, from the definition of evaluation dimensions to the generation of checklists. Section~\ref{sec:res} presents the experimental results with a comparative analysis of the tested models, while Section~\ref{sec:disc} discusses the main findings and limitations of RPA-Check and the performed experiments.  Finally, Section~\ref{sec:conc} concludes the paper, presenting future directions for research on the evaluation and application of LLM-based RPAs in specialized contexts.

\section{Related works}\label{sec:rel}

The integration of LLMs into gaming and simulation contexts is a rapidly evolving area of research characterized by multiple applications, ranging from the use of Generative AI as a development tool to its direct implementation in gameplay~\citep{Lanzi2023,Sweetser2024}. For example, recent works such as PLAYER~\citep{Zhu2025} explored multi-agent communication in murder mystery games, focusing on strategic reasoning and information hiding. Similarly, PsyPlay~\citep{Yang2025} introduced personality-infused role-playing for LLM agents, ensuring that these agents adhere to the so-called Big Five personality traits. The existing literature shows a growing interest in the use of LLMs to generate dynamic content~\citep{Farrokhi2024}, enable more natural interactions with NPCs~\citep{Cox2023}, and support immersive educational experiences~\citep{Song2024}. However, most approaches focus on game agents, content generation or dialogic interaction as an end in itself~\citep{Zhu2023,Ashby2023}, neglecting the need for rigorous methodologies to evaluate the quality and consistency of agentive behaviors in highly constrained and regulated scenarios~\citep{Wu2025,Sweetser2024}, such as the one chosen for this study i.e., simulated forensic settings.

To better highlight this multifaceted landscape and how the proposed contribution advances the state of the art, this section is organized into three parts. First, the use of Generative AI as a tool for content production and in-game integration, being the main application of LLM in video games, is examined (Section~\ref{subsec:rel:genAI}). Second, applications in educational and forensic simulation contexts, the validation domain of this work, are analyzed (Section~\ref{subsec:rel:sim}). Finally, existing methodologies for evaluating LLM-based agents and dialogic systems are presented (Section~\ref{subsec:rel:eval}).

\subsection{Generative AI in Games}\label{subsec:rel:genAI}

In the domain of content generation, numerous studies have explored how LLMs can automate the creation of narrative assets, game levels, and characters. \cite{Alavi2024} demonstrate the usefulness of LLMs as assistants in game plot design. Similarly, \cite{Kumaran2023} propose automating the generation of interactive narrative scenes using LLMs. While innovative, these approaches remain focused on the production phase and do not address the challenge of systematically evaluating the narrative quality or behavioral consistency of the generated agents. In contrast, the framework proposed in this research is not limited to generating content, but introduces a multi-stage evaluation pipeline that transforms qualitative role requirements into verifiable Boolean metrics, ensuring objective control over the fidelity of agent behaviors. \cite{Gursesli2023} use ChatGPT to generate visual novel narratives on climate themes. They actually propose to use ChatGPT as an automatic evaluator for the story quality by combining different prompts, but without systematically structuring granular questions for the evaluation criteria as in the approach proposed in this paper.

Games such as AI Dungeon~\citep{Hua2020} and Infinite Craft~\citep{Agarwal2024} integrate LLMs directly into the game loop, enabling procedural generation of content and responses in real time. \cite{Trukhin2026} explore the integration of LLM-powered NPCs in The Elder Scrolls V: Skyrim, finding an increase in perceived immersion among players. However, these implementations present limited control over long-term narrative coherence and the absence of clearly defined procedural constraints reduce their suitability in contexts where strict compliance with specific rules is essential. This work, on the other hand, introduces a forensic simulation environment in which agents must comply with explicit legal and procedural constraints, and proposes an automated evaluation mechanism capable of verifying such adherence through granular checklists generated and filtered algorithmically.

\subsection{Educational and forensic simulations}\label{subsec:rel:sim}

In the educational context, serious games have integrated Generative AI to improve teaching effectiveness through personalization and content adaptation. \cite{Zhao2024} present Language “Urban Odyssey”, a serious game for language learning that uses LLM to populate interactive environments with NPCs who speak a foreign language. \cite{Todova2025} proposes “A Quest for Information”, demonstrating that LLM-generated dialogues can increase engagement during educational events. However, these works do not address the issue of validating agent behaviors with respect to specific educational objectives, nor do they provide replicable methodologies for evaluating the pedagogical quality of the interaction. The framework proposed here fills this gap by introducing explicit evaluation dimensions (such as “Role Adherence” and “Logical Consistency”) and adopting an LLM-as-a-Judge approach to objectively quantify agents' adherence to assigned roles, enabling systematic comparative analysis between different models.

The forensic and investigative domain also presents a challenging terrain for LLM applications, given that it might require the management of dialogues constrained by specific (and coded) procedural rules and the ability to maintain logical consistency over extended sessions. Games such as “Verbal Verdict”~\citep{Savannadev2024} integrate LLMs to enable open-ended interrogations, but might suffer from hallucination and narrative inconsistencies. \cite{Beattie2017} discussed the potential of legal simulations games as a support for subjects not represented in courts, highlighting the need for procedural fidelity. Despite these efforts, to the best of our knowledge, the existing literature does not offer standardized methodologies for measuring the quality of forensic simulations or tools for objectively comparing the performance of different models in such simulation scenarios. This work tries to fill this gap by proposing an automated evaluation pipeline that adapts the CheckEval paradigm generating granual Boolean indicator checklists and applying chain-of-thoughts techniques to ensure reproducible verifications. The forensic and legal domain, with “LLM Court”, is the validation domain for our evaluation framework.

\subsection{Evaluation of LLM-based Agents in Video Games and Simulations}\label{subsec:rel:eval}

The scientific literature is rich in contributions to the evaluation of LLMs in general. SummEval~\citep{Fabbri2021} is a well-established framework for evaluating automatic summarization systems, based on various qualitative dimensions such as coherence, consistency, fluency, and relevance, measured through automatic metrics and human judgments. TopicalChat~\citep{Gopalakrishnan2023}, on the other hand, is a dataset and framework designed to evaluate knowledge-oriented open-domain dialogue systems, focusing on their ability to maintain natural and informative conversations on a variety of topics. CheckEval~\citep{Pereira2024} introduces the LLM-as-a-Judge paradigm, transforming qualitative evaluation dimensions into granular Boolean question checklists through an automated augmentation and filtering process, ensuring greater consistency and traceability compared to subjective human evaluations. However, although these approaches represent benchmarks in general, they were not designed to measure procedural fidelity and logical consistency in highly specialized contexts bound by explicit rules. SummEval and TopicalChat operate on general dimensions (textual coherence, informational relevance, conversational naturalness) which, while fundamental for assessing linguistic quality, do not capture critical aspects such as adherence to specific professional roles, compliance with strict procedural contraints, or the ability to maintain narrative coherence in structured multi-turn dialogues. Similarly CheckEval, while providing a robust methodological mechanism, has been validated primarily on summarization tasks and not on complex role-playing scenarios.

RPA-Check framework extends the CheckEval paradigm by tailoring a four-stage pipeline (Dimension Definition, Augmentation, Filtering, LLM-as-a-Judge Evaluation) to Role-Playing Agents. The proposed framework derives domain-specific evaluation dimensions (“Role Adherence”, “Argumentative Depth”, “Factual Consistency”, “Contextual Relevance”) by using, for validation, forensic and legal domain requirements. The framework allows generating checklists of Boolean indicators that query granular aspects such as the ability of agents to avoid narrative contradictions, respect procedural hierarchies (e.g., the order of speech between judge, prosecution, and defense), and maintain fidelity to the assigned role over prolonged dialogue sessions. Furthermore, while CheckEval has been applied mainly to single text generation tasks, this work adapts it to a dynamic multi-agent context, where evaluation must take into account the interactions between different agents and the overall consistency of the dialogic system.

Some recent benchmarks have expanded LLM-agent evaluation in gaming contexts. For example, TEXTQUESTS \citep{Phan2025} assesses long-horizon reasoning in 25 classic Infocom interactive fiction games requiring sustained problem-solving (such as “Zork”) over 100K+ token contexts without external scaffolding. FAIRGAMER~\citep{Shi2026} instead evaluates social biases in LLM-based NPCs through game-theoretic interaction patterns (transaction, cooperation, competition) across 16,910 bilingual test cases. While these works address long-context reasoning and fairness respectively, neither tackles the core challenge of constraint adherence in multi-agent role-playing, which is, on the contrary, the focus of LLM Court. TEXTQUESTS evaluates single-agent exploratory scenarios, whereas our forensic simulation requires simultaneous satisfaction of role-specific behavioral norms, logical consistency, and procedural compliance across multi-turn interactions. FAIRGAMER's fairness metrics are orthogonal to our evaluation objectives: LLM Court prioritizes behavioral fidelity to domain-specific constraints rather than equitable decision-making. Similarly to TEXTQUESTS and FAIRGAMER, ORAK~\citep{Park2025} proposes a large-scale benchmark spanning 12 commercial video games across multiple genres, supporting agentic modules such as reflection and planning and evaluating models through gameplay performance metrics and leaderboards. While ORAK emphasizes cross-genre generalization and strategic competence in dynamic environments, its evaluation remains performance-driven, with success defined by score maximization or task completion. In contrast, LLM Court evaluates agents in a tightly regulated forensic setting where correctness is defined by sustained compliance with explicit procedural and role-specific constraints. Thus, whereas ORAK measures gameplay effectiveness, LLM Court measures institutional fidelity under structured normative requirements.

In summary, although the existing literature has explored the integration of LLMs in video games and serious games from multiple perspectives, significant gaps remain regarding the systematic evaluation of agents' fidelity to complex and constrained roles. The RPA-Check framework proposed in this work fills this gap by providing a rigorous, automated, and replicable methodology for evaluating RPAs in forensic contexts, empirically demonstrating how small, appropriately evaluated models can outperform larger models in terms of narrative consistency and stability.

\section{Methodology: The Multi-Stage Automated Evaluation Framework}
\label{sec:method}

Evaluating RPAs in constraints-heavy environments presents a unique challenge: standard NLP metrics (e.g., BLEU, ROUGE, Perplexity) fail to capture high-level semantic requirements like role adherence, narrative stability, or procedural compliance. A model might generate linguistically fluent text that is nonetheless procedurally invalid or logically incoherent.
To address these limitations, we propose RPA-Check, a multi-stage automated evaluation framework that transforms qualitative behavioral requirements into granular, verifiable boolean indicators via an ``LLM-as-a-Judge'' approach for the final evaluation.

\subsection{RPA-Check Framework Architecture}
\begin{figure*}
    \centering
    \includegraphics[width=0.99\textwidth]{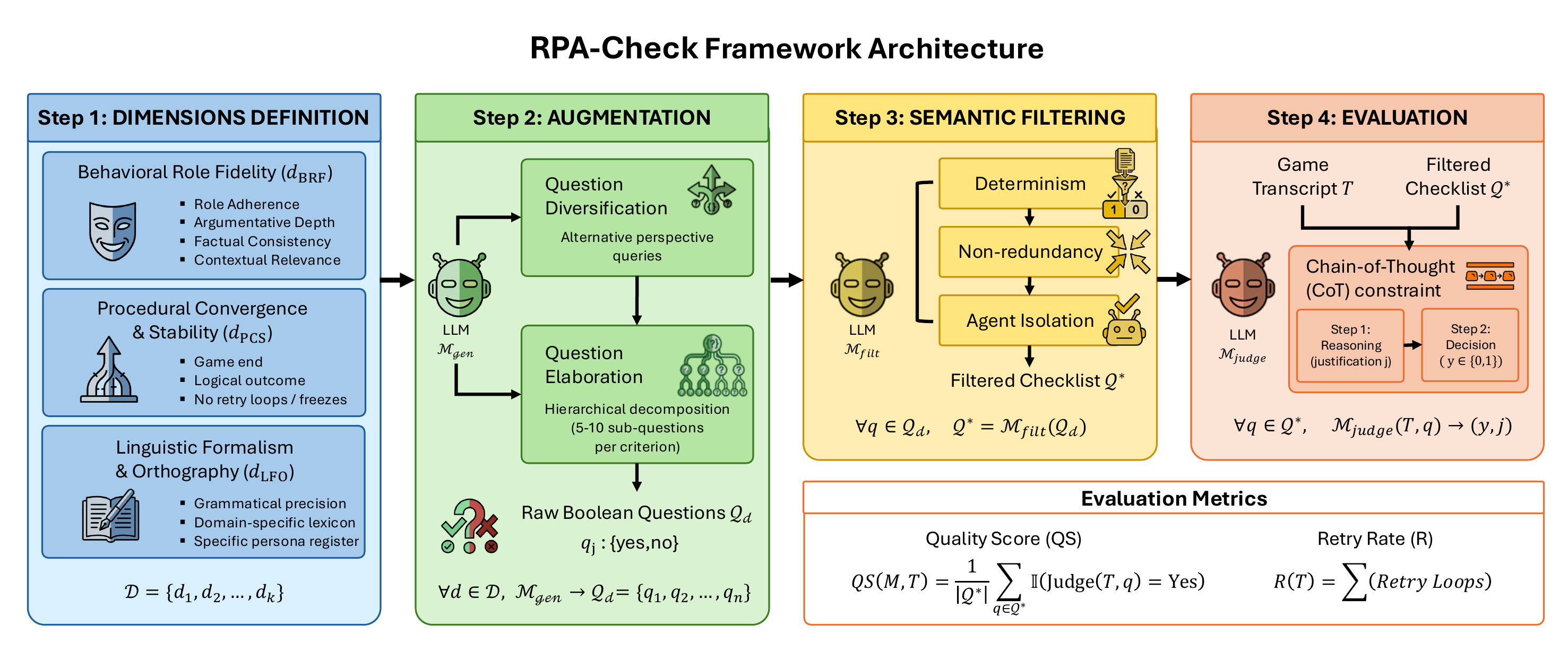}
    \caption{Architecture of the RPA-Check Framework. The automated evaluation pipeline transforms qualitative behavioral requirements of Role-Playing agents into quantitative metrics through a four-stage process: \textit{Dimensions Definition} establishes behavioral, procedural, and linguistic criteria; \textit{Augmentation} expands these dimensions into granular Boolean checklists via an LLM generator; \textit{Semantic Filtering} ensures checklist objectivity and agent isolation; and \textit{Evaluation} based on LLM-as-a-Judge approach scores trial transcripts using a Chain-of-Thought (CoT) reasoning protocol. Final performance is quantified via the Quality Score (QS) and Retry Rate (R) metrics to assess role fidelity and operational stability.}\label{fig:framework}
\end{figure*}

The framework is structured as a four-stage pipeline designed to minimize evaluator subjectivity and maximize the resolution of the performance metrics (Figure \ref{fig:framework}).

\paragraph{Stage 1: Dimension Definition}
The first stage of the framework involves the formal definition of a set $\mathcal{D}$ of ${K}$ high-level evaluation dimensions, denoted as $\mathcal{D} = \{d_1, d_2, \dots, d_K\}$, where ${K}$ represents the total number of macro-criteria selected for the benchmark. Each dimension  $d_i \in \mathcal{D}$ encapsulates a distinct qualitative behavior attribute of the agents, serving as the basis for the subsequent automated generation of granular indicators.

To ensure a comprehensive assessment of behavioral fidelity, system stability and quality, for our specific implementation, we define $K=3$ primary macro-dimensions:

\begin{itemize}
    \item \textbf{Behavioral Role Fidelity ($d_{BRF}$):} 
    this dimension serves as the primary metric for validating agent personas. It evaluates the degree to which an agent maintains functional boundaries and avoids sycophancy. It is treated as a composite vector comprising four core sub-dimensions:
    \begin{itemize}
        \item \textit{Role Adherence:} the agent must operate strictly within its assigned role $\pi_{role}$ while suppressing capabilities of other roles.
        \item \textit{Argumentative Depth:} it measures the semantic complexity of responses, awarding high scores for multi-step logical arguments and referencing specific evidence.
        \item \textit{Factual Consistency:} it verifies that assertions do not contradict previous turns or the immutable ground truth established in the case file.
        \item \textit{Contextual Relevance:} it assesses the causal link between consecutive dialogue turns, ensuring the response directly addresses the query or stimulus provided in the previous turn. It specifically penalizes "non-sequiturs" where the agent ignores a direct question.        
    \end{itemize}

    \item \textbf{Procedural Convergence and Stability ($d_{PCS}$):}
    this dimension assesses the ability of the system to drive the narrative toward a logical and successful conclusion. For instance, in the context of the \textit{LLM Court} framework applied in this study, it measures the adherence to the deterministic Finite State Machine (FSM) governing the simulation. A trial is considered procedurally stable when: i) the interaction successfully reaches the end of the game; ii) the final outcome is semantically and logically aligned with the evidence accumulated during the dialogue history; iii) the simulation completes without requiring manual ``Retry'' loops or encountering catastrophic model freezes.

    \item \textbf{Linguistic Formalism and Orthography ($d_{LFO}$):} 
    linguistic fidelity is critical for maintaining the immersion and professional register of the RPA required in specialized domains. This dimension ensures that the generated text adheres to grammatical precision, domain-specific lexicon and specific persona register.

\end{itemize}

\paragraph{Stage 2: Augmentation} 
A generator model $\mathcal{M}_{gen}$ is employed to enrich each dimension $d_i$ into a set of granular Boolean questions $\mathcal{Q}_d$. This stage utilizes two distinct prompting strategies:
\begin{enumerate}
    \item \textbf{Question Diversification:} generates alternative queries to analyze the same concept from multiple perspectives.
    \item \textbf{Question Elaboration:} decomposes main criteria into a hierarchy of sub-questions (5 to 10 per criterion) to achieve a fine-grained analysis.
\end{enumerate}

Formally, for any dimension $d_i \in \mathcal{D}$:
\begin{equation}
    d \xrightarrow{\mathcal{M}_{gen}} \mathcal{Q}_d = \{q_1, q_2, \dots, q_n\}
    \label{eq:generator}
\end{equation}
where each $q_j$ must be answerable with a binary $\{yes, no\}$.
This step increases the resolution of the evaluation, minimizing the variance typically associated with single-score Likert scales.

\paragraph{Stage 3: Semantic Filtering} 
The raw set $\mathcal{Q}_d$ undergoes a filtering process to ensure objectivity and relevance. This stage is performed by a specialized filtering model $\mathcal{M}_{filt}$, which is tasked with removing redundant, subjective, or non-pertinent indicators to produce the final checklist $\mathcal{Q}^*$. Formally, the filtering operation is defined as:
\begin{equation}
\mathcal{Q}^* = \mathcal{M}_{filt}(\mathcal{Q}_d)
\label{eq:filter}
\end{equation}
where $\mathcal{Q}^*$ denotes the subset of indicators for all questions dimension $\mathcal{Q}_d$ that satisfy the following operational constraints:

\begin{itemize}
    \item Binary determinism: each question must be answerable with a strictly binary {yes,no} value based exclusively on the interaction transcript T.
    \item Non-redundancy: overlapping or duplicate questions are identified and removed to ensure each indicator provides a unique informative contribution.
    \item Agent isolation: questions concerning human-controlled participants are discarded to isolate the performance of the generative models.
\end{itemize}

\paragraph{Stage 4: LLM-as-a-Judge Evaluation} 
The final evaluation is performed by a high-parameter Judge Model ($\mathcal{M}_{judge}$). For a transcript $T$ and the filtered checklist $\mathcal{Q}^*$, the judge generates a binary decision $y \in \{0, 1\}$ and a justification $j$ for each $q \in \mathcal{Q}^*$. We enforce a \textit{Chain-of-Thought} (CoT) constraint to ground the evaluation:
\begin{equation}
    \forall q \in \mathcal{Q}^*, \quad \mathcal{M}_{judge}(T, q) \rightarrow (y, j)
    \label{eq:judge}
\end{equation}

\subsection{Evaluation Metrics}
\label{subsec:metrics}

To quantitatively assess the performance of the RPAs, we define two primary metrics that capture both the qualitative performance and the operational reliability of the models:

\begin{itemize}
    \item \textbf{Quality Score (QS):} this is the fundamental metric used to evaluate all three macro-dimensions $d_{BRF}$, $d_{PCS}$, and $d_{LFO}$. The $QS$ represents the percentage of affirmative responses provided by the Judge model across the filtered boolean checklists. For a given model $M$ and transcript $T$, it is formally defined as:
    \begin{equation}
        QS(M, T) = \frac{1}{|\mathcal{Q}^*|} \sum_{q \in \mathcal{Q}^*} \mathbb{I}(\text{Judge}(T, q) = \text{Yes})
        \label{eq:qs}
    \end{equation}
    where $\mathbb{I}(\cdot)$ is the indicator function. By applying $QS$ to all dimensions, we ensure a standardized measurement of qualitative attributes, ranging from persona adherence to the logical consistency of the final verdict.

    \item \textbf{Retry Rate ($R$):} this metric quantifies the stochastic instability and operational failures of the models. $R$ is defined as the total number of manual restarts required due to catastrophic failures during a simulation, such as infinite loops, model freezing, or the inability to correctly tag the next speaker. While $QS$ inherently measures even the quality of successful interactions, $R$ serves as the negative counterpart to stability: a higher $R$ indicates that a model is prone to narrative degradation regardless of its linguistic fluency.
\end{itemize}

\section{The LLM Court Case Study}\label{sec:sysDes}
To validate our framework, we implemented \textit{LLM Court}, a platform designed to test RPAs in a rules-based environment conceived to simulate forensic dialogue under strict procedural constraints. Unlike open-ended conversational agents utilized in purely recreational contexts, our system enforces a rigid separation between the generative logic (content production) and the procedural control (courtroom rules).
This section describes the system architecture used to generate the interaction data for our evaluation.

\subsection{Architectural Overview}

The architecture is built upon a local-inference engine integrated within the Unity environment, utilizing quantized LLMs to drive the NPCs. This architectural choice ensures operational independence and data privacy, a critical requirement for potential educational or professional applications in legal training. The game core loop consists of three distinct computational modules:
\begin{itemize}
    \item \textit{Procedural Case Generation:} it instantiates the semantic boundaries of the simulation via remote high-parameter models.
    \item \textit{Agentic State Machine:} a deterministic finite-state automation governing turn-taking dynamics between the Player (Defense) and RPAs (Judge, Prosecutor, Witness). The dynamic role-playing of the agents is regulated via system prompt.
    \item \textit{Intent Analysis Module:} a parallel processing unit responsible for extracting structured metadata from unstructured natural language outputs.
\end{itemize}

\begin{figure*}
    \centering
    \begin{subfigure}{0.49\textwidth}
        \centering
        \includegraphics[width=\textwidth]{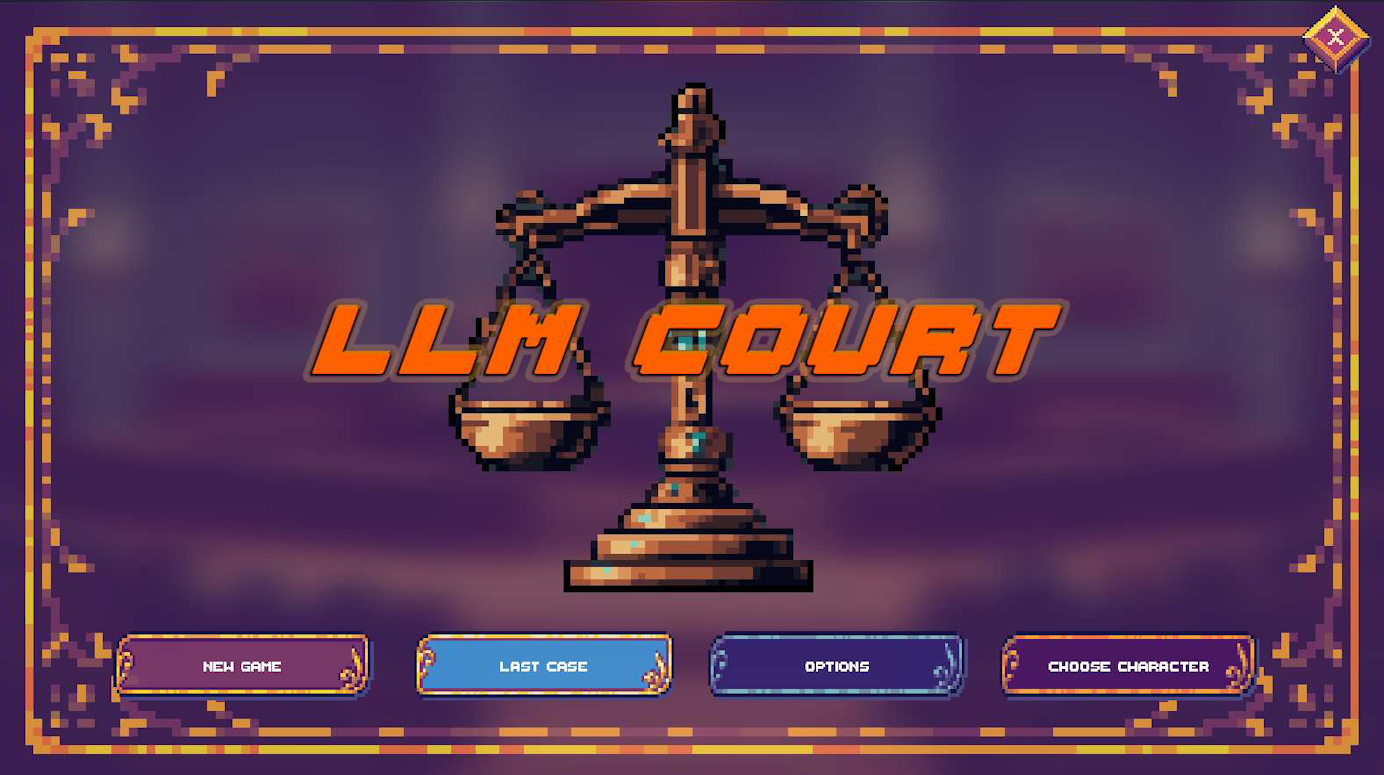}
        \caption{}\label{subfig:ui}
    \end{subfigure}
    ~
    \begin{subfigure}{0.49\textwidth}
        \centering
        \includegraphics[width=\textwidth]{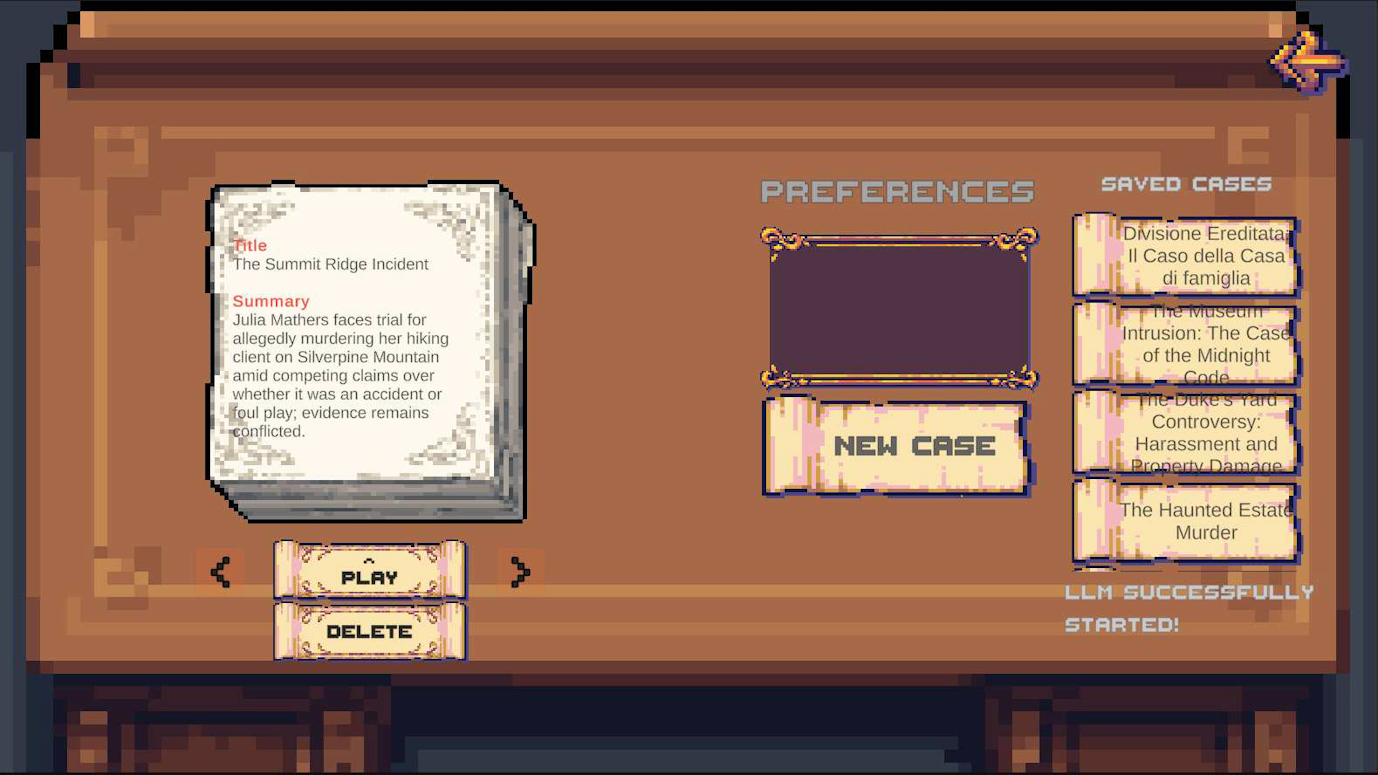}
        \caption{}\label{subfig:case-gen}
    \end{subfigure}
    \\
    \begin{subfigure}{0.49\textwidth}
        \centering
        \includegraphics[width=\textwidth]{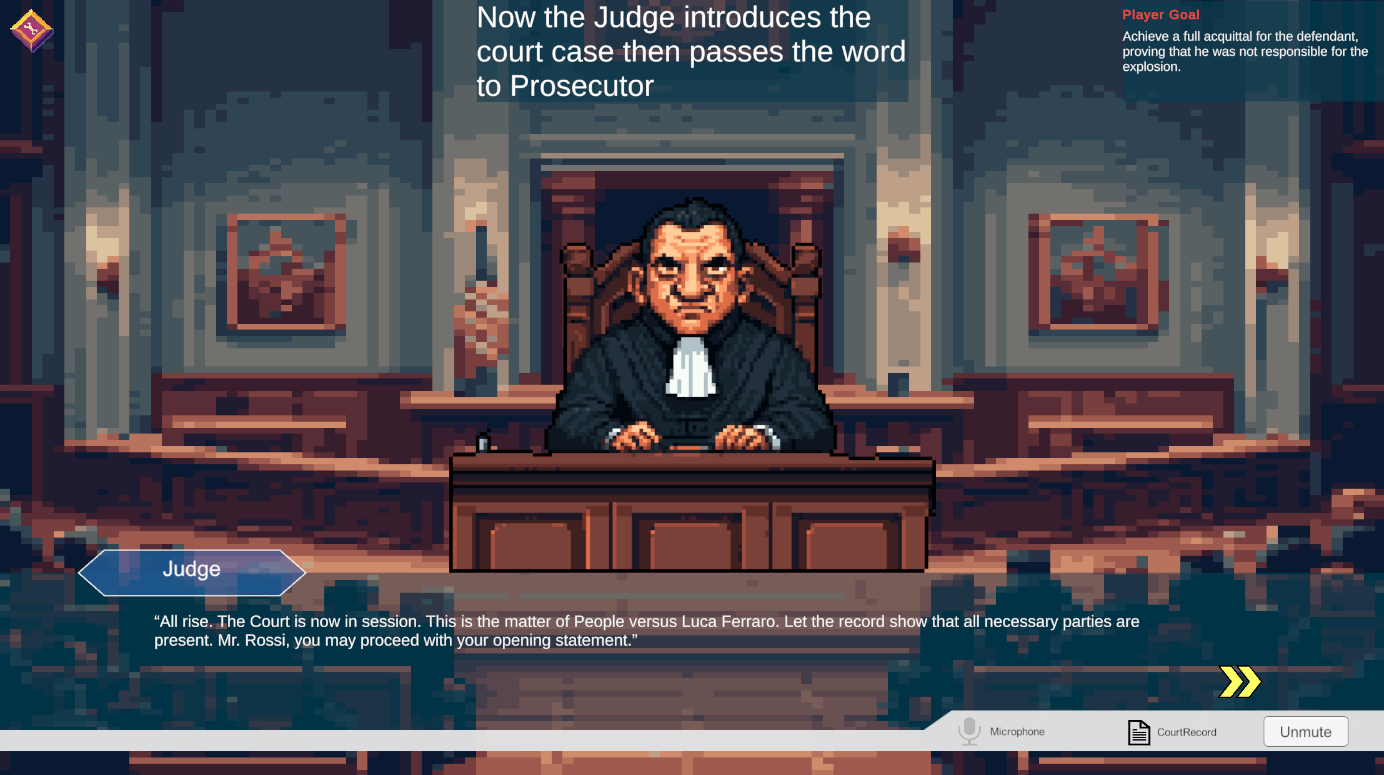}
        \caption{}\label{subfig:judge}
    \end{subfigure}
    ~
    \begin{subfigure}{0.49\textwidth}
        \centering
        \includegraphics[width=\textwidth]{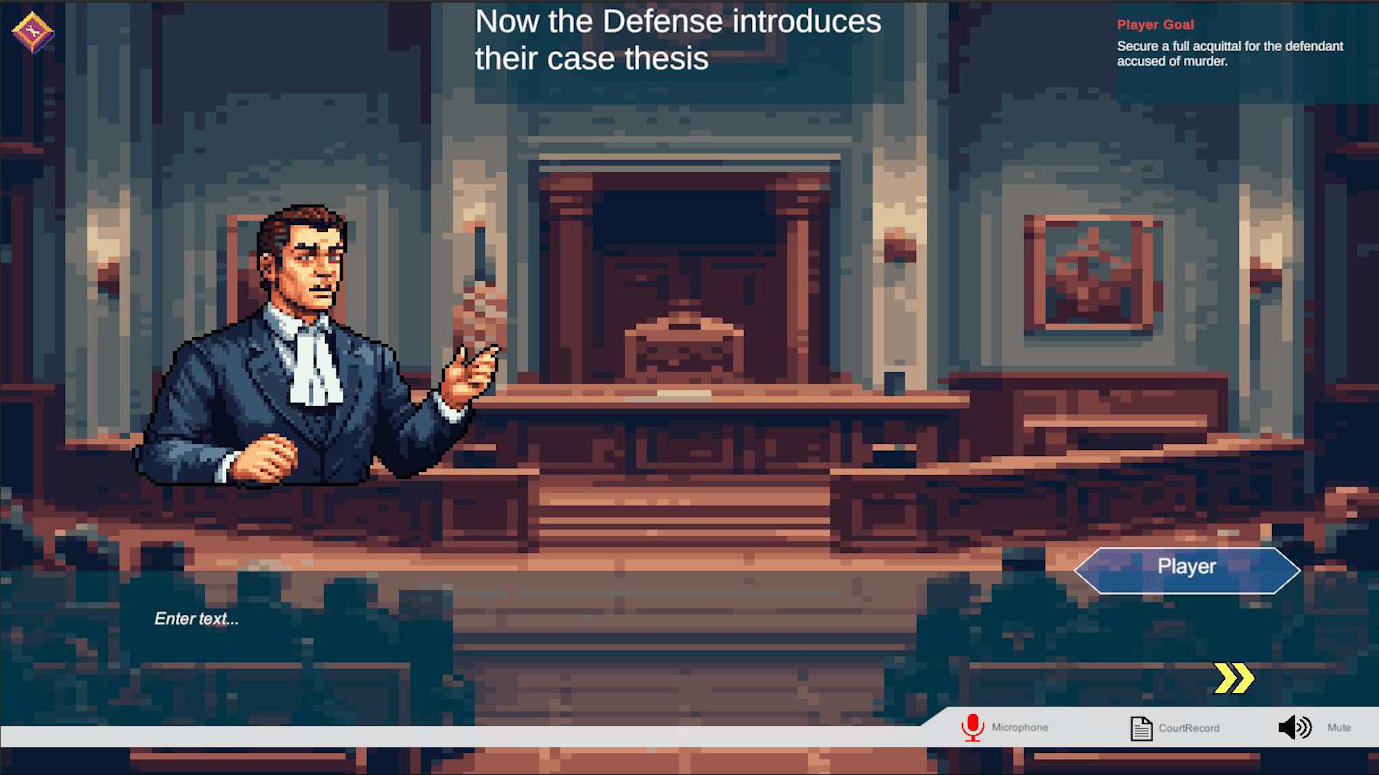}
        \caption{}\label{subfig:player}
    \end{subfigure}
    \caption{Screens from LLM Court: (a) the main game menu and (b) the case generation module. During the simulation, the game transitions to the courtroom, where the judge RPA (c) introduces the case, and the player (d) interacts with the RPAs through text-based input.}\label{fig:game}
\end{figure*}

\begin{figure*}
    \centering
    \includegraphics[width=0.99\textwidth]{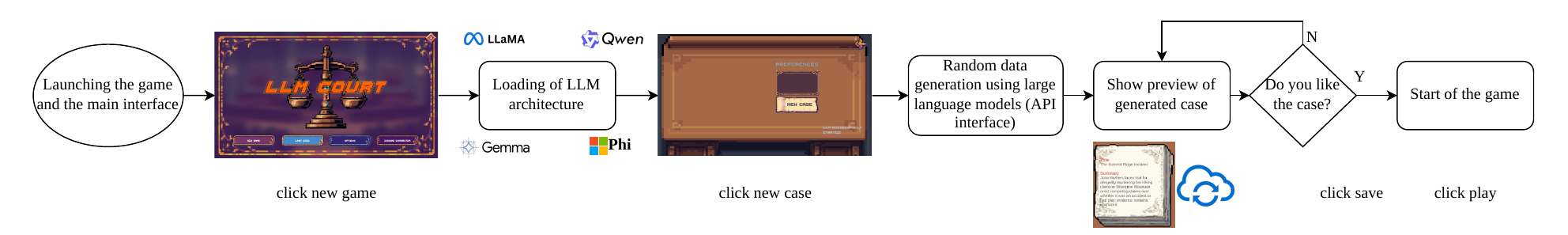}
    \caption{The player's flow in LLM Court. After launching the game interface, starting a new game loads the RPAs LLM. Upon selecting a “new case,” the Procedural Case Generation module powered by GPT-5 generates and displays a case description. The player may either save the generated case to begin the game or discard it and request a new one.}\label{fig:user-flow}
\end{figure*}

Figure~\ref{fig:game} shows some screens from LLM Court, such as the main game menu where the player starts a new game or loads a previous one, sets the game's options and chooses their avatar (\ref{subfig:ui}), and the user interface for Procedural Cage Generation (\ref{subfig:case-gen}). During the simulation, the game is set in a courtroom, where the judge RPA (\ref{subfig:judge} introduces the case and the player (\ref{subfig:player}) interacts with the LLM Court RPAs via text-based input.
Specifically, the player's flow in LLM Court is schematized in Figure~\ref{fig:user-flow}: after launching the game and its main interface, starting a new game loads the RPAs LLM. Following the player selection of a “new case”, the Procedural Case Generation is performed through GPT-5, showing the generated text. If the player saves the case, the game actually starts otherwise a new case is generated.

\subsection{Procedural Case Generation via Structured Sampling}
\label{sec:case_gen}
To ensure narrative variety and logical consistency, the initialization phase employs a stochastic generation process constrained by a predefined schema. While run-time interaction relies on quantized local models for efficiency, we utilize GPT-5 via Azure OpenAI Service exclusively for this pre-computation phase to maximize coherence in the case facts.

Formally, a legal case $\mathcal{C}$ is defined as a tuple $\mathcal{C} = \{T, S, \mathcal{E}, \mathcal{W}, G\}$, where $T$ is the title, $S$ is the factual summary, $\mathcal{E}$ is the set of evidentiary items, $\mathcal{W}$ is the set of witness profiles, and $G$ denotes the defense goal. The generation process minimizes the probability of hallucination by strictly enforcing a JSON output format. Let $P_{\theta}$ be the language model parameterized by $\theta$. The generation of $\mathcal{C}$ is conditional on the schema constraints $\mathcal{I}_{schema}$ and user-defined priors $\mathcal{P}_{user}$ (e.g., "theft case"):

\begin{equation}
    \mathcal{C} \sim P_{\theta}(\mathcal{C} \mid \mathcal{I}_{schema}, \mathcal{P}_{user})
\end{equation}

This module deserializes the output into run-time objects, populating the ``Court Record,'' which serves as the immutable ground-truth knowledge base. This explicitly prevents the \textit{dynamic} agents from hallucinating facts that contradict the \textit{static} case file during the trial phase.

\subsection{Dynamic Role-Playing via System Prompt Injection}
The core interaction is driven by a dynamic prompting strategy that adapts the model’s persona to the current state of the trial. We implement a context injection mechanism that prepends a composite System Prompt $\mathcal{S}_{role}$ to the dialogue history $H_t$ at turn $t$. To ensure complete session independence and introduce variability, given that such variability can be obtained with a small change in input~\citep{Salinas2024,Sclar2024}, each simulation defines a generation seed included in the agents' initial prompt (i.e., an integer parameter). Varying the seed across repeated runs of an identical scenario produces distinct dialogue trajectories while holding all other experimental variables constant.
The System Prompt is engineered to mitigate two pervasive issues identified in LLM-based role-playing literature: i) Recency Bias (Instrunction Drift), the tendency of autoregressive models to prioritize recent dialogue tokens over initial behavioral instructions; ii) User-Alignment Bias (Sycophancy), the inherent tendency of instruction-tuned models to be cooperative towards the user, which is detrimental in an adversarial forensic setting.

To address these, we define distinct prompt architectures for the key roles:

\subsubsection{The Judge (Controller Agent)}
The Judge acts as a deterministic regulator rather than a conversationalist. The prompt enforces a ``Chain-of-Thought'' (CoT) logic to evaluate the admissibility of arguments before passing the turn. The Judge's policy $\pi_{judge}$ is defined not to maximize engagement, but to minimize procedural entropy:
\begin{equation}
    \pi_{judge}(a_t \mid H_t) \rightarrow \{ \text{Grant Intervention}, \text{Overrule}, \text{Sustain}, \text{Verdict} \}
\end{equation}

\subsubsection{The Prosecutor (Adversarial Agent)}
The Prosecutor agent is engineered to counteract the sycophancy alignment of standard base models.
We formalize the Prosecutor's objective as prioritizing adversarial argumentation over user alignment. Conceptually, this agent optimizes a policy $\pi_{pros}$ that maximizes the logical strength of the accusation $L_{acc}$ while penalizing semantic alignment with the user's intent $I_{user}$. The optimal response $x_t^*$ is derived as follows:

\begin{equation}
    \pi_{pros}(H_t) = x_t^* = \operatorname*{argmax}_{x \in \mathcal{V}} \left[ \underbrace{P_{\theta}(x \mid H_t, \mathcal{S}_{pros})}_{L_{acc}(x)} - \lambda \cdot \underbrace{\mathcal{A}(x, I_{user})}_{\text{Penalty}} \right]
\end{equation}

Where $\mathcal{S}_{pros}$ represents the prompt instructions explicitly forbidding neutral or helpful behaviors (e.g., \textit{"Do not help the Defense"}, \textit{``Focus on the weakness of the user's argument''}), $\mathcal{A}(x, I_{user})$ is an alignment function that measures the semantic similarity between the candidate response and the user's goal and $\lambda$ is the adversarial coefficient, penalizing responses that are helpful to the user.

\subsubsection{The Witness (Informational Agent)}
Unlike the Judge (who regulates) or the Prosecutor (who competes), the Witness functions as a constrained informational node. The primary design challenge for this role is balancing narrative realism with factual Consistency.
Standard LLMs tend to act as "perfect databases" or, conversely, hallucinate details to fill narrative gaps. To address this, the Witness system prompt imposes a bounded knowledge constraint. The agent is instructed to provide realistic testimony where human-like uncertainty is explicitly permitted (\textit{``Uncertainty is allowed, but responses must be logically coherent''}), yet strictly anchored to the immutable case generation described in Section \ref{sec:case_gen}.

Formally, let $K_{GT}$ be the Ground Truth knowledge contained in the case file, and $P_{pers}$ be the personality profile of the specific witness. The Witness policy $\pi_{wit}$ generates a response $x_t$ to a query $q_t$ by maximizing adherence to the personality while prohibiting information retrieval outside of $K_{GT}$:

\begin{equation}
    x_t \sim P_{\theta}(x_t \mid q_t, H_t, P_{pers}) \quad \text{s.t.} \quad \text{Consistency}(x_t, K_{GT}) = 1
\end{equation}

where the consistency function penalizes any assertion $a \in x_t$ that contradicts the set of evidentiary facts $\mathcal{E} \subset K_{GT}$. This explicitly prevents the "Creative Hallucination" phenomenon where witnesses might invent exculpatory or incriminating evidence that does not exist in the simulation's state space.

\subsection{Finite-State Turn Management via Sentence Analysis}
A critical innovation of \textit{LLM Court} is the abstraction of turn-taking from the language model’s latent space to a deterministic Finite State Machine (FSM). While the \textit{content} of a response is generative, the \textit{decision} of who speaks next is handled by the Sentence Analyzer.

This auxiliary component runs a parallel inference task on the generated output to extract routing tags (e.g., \texttt{<NextSpeaker: Witness\_A>}). This prevents the LLM from hallucinating procedural flows, such as a witness interrogating the judge, by enforcing a rigid transition graph $\mathcal{G} = (V, E)$, where $V$ represents the set of actors and $E$ the allowable transitions.

\begin{figure*}
    \centering
    \includegraphics[width=0.99\textwidth]{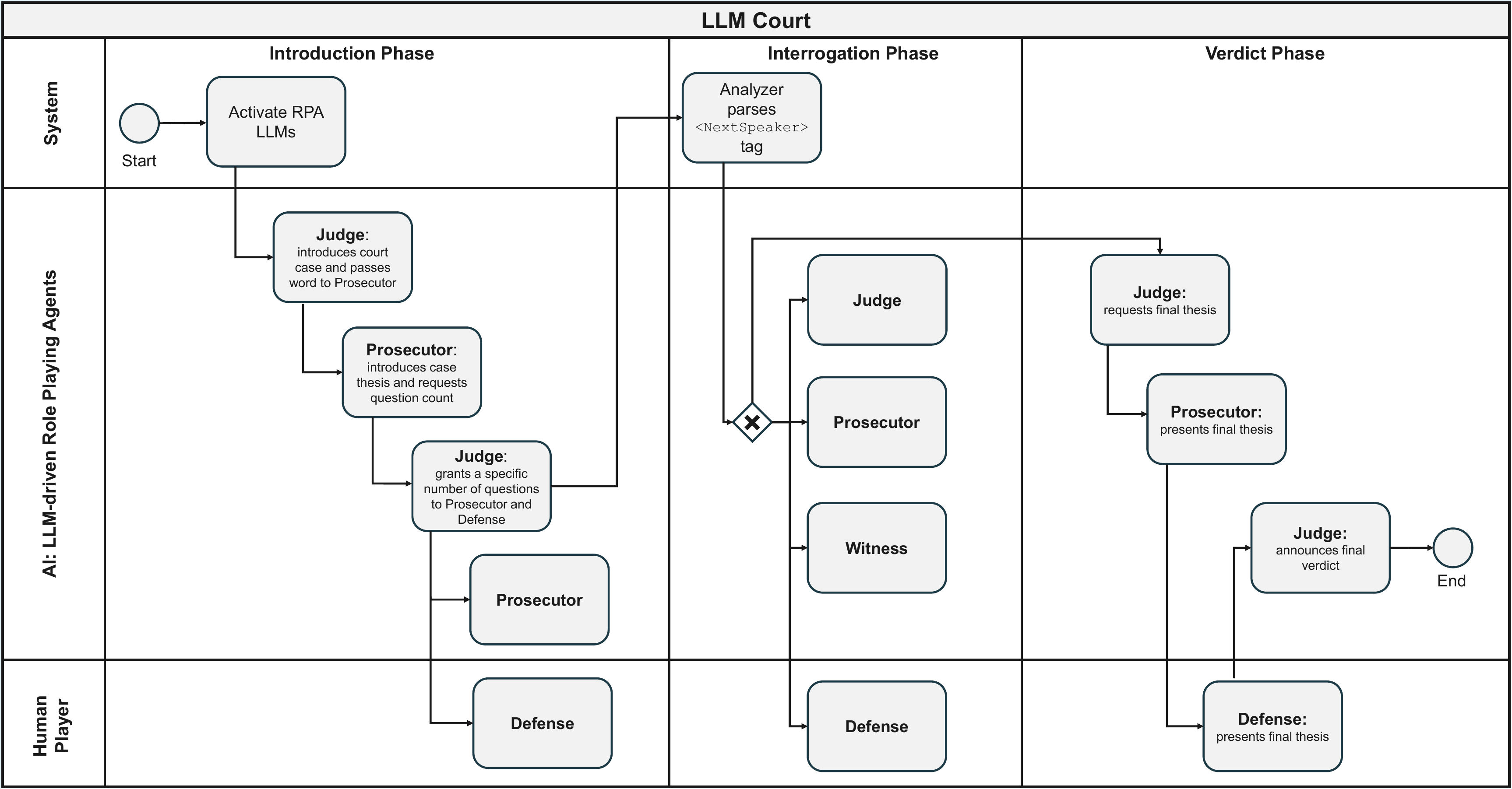}
    \caption{Schematic of the LLM Court simulation flow. The diagram illustrates the multi-phase execution logic managed by the FSM. The pools delineate the operational boundaries between the system-level procedural logic, the AI, i.e., the LLM-driven RPAs, and the interactive human player input (Defense). Process Logic is divided into three distinct chronological stages: the Introduction Phase (fixed sequential setup), the Interrogation Phase (dynamic dialogue), and the Verdict Phase (unscripted argumentation and final conclusion). In the dynamic dialogue, the Analyzer determines the turn-taking order at runtime via specific <NextSpeaker> tags, diverging from standard sequential flow to simulate realistic courtroom interactions.}\label{fig:game-flow}
\end{figure*}

The simulation flow (Figure~\ref{fig:game-flow}) is segmented into three phases:
\begin{enumerate}
    \item Introduction Phase: fixed sequence ($Judge \rightarrow Prosecutor \rightarrow Defense$).
    \item Interrogation Phase: dynamic branching. The Analyzer parses the user's input to identify the addressee $y \in \mathcal{W}$.
    \item Verdict Phase: synthesis and evaluation.
\end{enumerate}

\subsection{Verdict Logic}
To ensure the simulation concludes with a logically sound verdict, we implement a Summarization-Evaluation Pipeline. Upon the termination of the dialogue rounds, the system triggers a background process.

Let $H_{final}$ be the complete dialogue history. The process involves two steps:
\begin{enumerate}
    \item Summarization: an abstraction function $f_{sum}$ compresses the history into a salient summary $A$:
    \begin{equation}
        A = f_{sum}(H_{final})
    \end{equation}
    \item Evaluation: a separate instance of the model, instantiated with the Judge's rubric $R_{eval}$, maps the summary and the initial goal $G$ to a binary outcome $O \in \{0, 1\}$ (Loss/Win) and a justification $J$:
    \begin{equation}
        (O, J) \sim P_{\theta}(O, J \mid A, G, R_{eval})
    \end{equation}
\end{enumerate}

This separation of concerns reduces the cognitive load on the model during the trial and ensures that the final verdict is derived from the accumulated evidence rather than the sentiment of the most recent tokens.

\section{Experimental Evaluation}\label{sec:res}
The experiments applied RPA-Check, to the procedural simulations generated by LLM Court. After presenting the evaluation setup~\ref{subsec:res:setup}, the results, presented in Subsection~\ref{subsec:res:res} are organized into two levels of granularity: an aggregated analysis over all the legal scenarios and an analysis by individual scenario, in order to highlight both average performance and behavioral variance, both of which are relevant dimensions in the LLM court evaluation and to demonstrate the proposed framework.

\subsection{Experimental Setup}
\label{subsec:res:setup}
The proposed framework was validated by applying it to the LLM Court  case study (described in Section \ref{sec:sysDes}). Through the automated augmentation and semantic filtering pipeline, we derived a final evaluation checklist consisting of 38 granular boolean questions.
The framework's efficacy was validated through an ablation study across varying model architectures and sizes.

\paragraph{Models Evaluated} 
We evaluated seven open-weight models categorized by parameter count into Small/Efficient models (Llama-3.1-8B, Gemma-2-9B-it, Qwen-3-8B, and Hermes-3-8B) and Medium/Large (Gemma-3-12B-it, Qwen-3-14B, and Phi-4-14B).

\paragraph{Quantization and Deployment} 
To ensure reproducible results in resource-constrained environments, all models were 4-bit or 5-bit quantized (GGUF) and executed locally via the Unity environment using the \texttt{LLMUnity} library.

\paragraph{Legal Scenario Corpus} 
We generated five distinct procedural legal scenarios, ranging in complexity from theft to murder, using GPT-5 via Azure OpenAI Service to serve as the ground-truth context. The scenarios are as follows:
\begin{itemize}
    \item \textit{Case 1}: the Summit Ridge Incident. Mountain guide Julia Mathers is charged with murdering a client found dead near a perilous ledge on Silverpine Mountain. Ambiguous physical evidence, conflicting witness timelines, and contested motive make this scenario particularly effective for evaluating factual consistency and the agents' resistance to speculative reasoning.
    \item \textit{Case 2}: the Case of the Château Trespass. Luc Morel is accused of unlawfully entering part of a historic French château estate, with the dispute centering on ambiguous property boundaries and disputed prior permissions. This scenario tests contextual relevance and the agents' capacity to maintain legally nuanced argumentation under evidentiary ambiguity.
    \item \textit{Case 3}: the Museum Intrusion: The Case of the Midnight Code. Software developer Alex Morgan faces charges of artifact theft and unauthorized computer access at a city museum, with digital evidence suggesting potential framing. The technically complex evidence chain challenges agents' factual consistency and their ability to reason coherently across multi-turn interrogations.
    \item \textit{Case 4}: the Duke's Yard Controversy: Harassment and Property Damage. A civil dispute in which Harold Grayson is accused of harassment and intentional destruction of bamboo structures on a neighboring estate. The contested boundary evidence and witness bias render this scenario well-suited for assessing role adherence in non-criminal, adversarial civil proceedings.
    \item \textit{Case 5}: the Haunted Estate Murder. Former groundskeeper Marcus Graves is accused of murdering heiress Vivian Blackthorn at a Halloween costume party, amid circumstantial physical evidence and unreliable witness accounts. The multiplicity of witnesses and the inherently ambiguous testimony make this scenario the most demanding in terms of argumentative depth and long-term narrative stability.
\end{itemize}

\paragraph{Simulation Protocol} 
During each trial, the model under evaluation controlled all NPC roles (Judge, Prosecutor, and Witnesses). Crucially, the Defense Attorney role was played by the human player, providing the necessary ``adversarial'' input to drive the simulation.

\paragraph{Automated Judge} 
The final evaluation was performed by an ``LLM-as-a-Judge'' (i.e. GPT-5) processing the full interaction transcripts. To isolate the performance of the agents, any interventions by the human player were explicitly excluded from the scoring process. The judge followed a \textit{Chain-of-Thought} (CoT) constraint, providing a brief justification in parentheses after each binary answer to mitigate hallucinations and ground the evaluation in specific dialogue turns.

\subsection{Experimental Results}\label{subsec:res:res}
Table~\ref{tab:res:agg} presents an aggregate analysis of the Behavioural Role Fidelity Quality Score ($QS_{d_{BRF}}$). It implicitly defines three groups of models. The first, consisting of Llama-3.1-8B, Gemma-2-9B-it, and Qwen-3-8B, has an average $QS_{d_{BRF}}$ between 0.89 and 0.92, placing it above the sample's average threshold. The second group includes Gemma-3-12B-it and Hermes-3-8B, with scores of 0.82 and 0.81 respectively. The third group, consisting of Phi-4-14B and Qwen-3-14B, has the lowest values, 0.76 and 0.73, despite their larger parameter sizes. This distribution suggests that the scale is not a reliable predictor of performance in highly procedurally constrained contexts.
\begin{table}
\centering
\caption{Aggregate performance of models across the five legal scenarios. The $QS_{d_{BRF}}$ represents the average Quality Score (QS) of the Behavioural Role Fidelity dimension on successfully completed cases. The Retry Rate ($R$) indicates the total number of restarts requested.} \label{tab:res:agg}
\begin{tabular}{lccc}
\hline
\textbf{Model}  & \textbf{Quantization} & \textbf{Retry Rate ($R$)} & \textbf{$QS_{d_{BRF}}$} \\
\hline
Llama-3.1-8B   & 4-bit & 3 & 0.925 \\
Gemma-2-9B-it  & 4-bit & 0 & 0.893 \\
Qwen-3-8B      & 4-bit & 1 & 0.893 \\
Gemma-3-12B-it & 4-bit & 0 & 0.823 \\
Hermes-3-8B    & 5-bit & 3 & 0.810 \\
Phi-4-14B      & 4-bit & 0 & 0.760 \\
Qwen-3-14B     & 4-bit & 0 & 0.730 \\
\hline
\end{tabular}
\end{table}

Table~\ref{tab:res:agg:subdim} breaks down the results along the axes of Role Adherence (D1), Argumentative Depth (D2), Factual Consistency (D3), and Contextual Relevance (D4), allowing us to identify specific patterns of failure or excellence for each architecture. These results exhibit a systematic trade-off between D2 and D1 that is particularly pronounced in larger models. Qwen-3-14B achieves maximum score on D2 but drops to 0.46 on D1, indicating that the generated responses have high argumentative complexity but frequent deviations from the assigned role. Gemma-3-12B-it shows a similar profile, with 1.00 on D2 and 0.73 on D1. In contrast, the Llama-3.1-8B, Gemma-2-9B-it, and Qwen-3-8B models maintain high values on D4 (Contextual Relevance, 1.00), confirming a greater ability to preserve the sequential consistency of the dialogue. The D3 sub-dimension (Factual Consistency) is the most variable among the models, with scores ranging from 0.63 to 0.8, indicating a vulnerability in generating assertions that are inconsistent with the court case provided as ground truth.
\begin{table}
\centering
\caption{Breakdown of the Behavioural Role Fidelity Quality Score ($QS_{d_{BRF}}$) for its four sub-dimensions. The given percentage of affirmative responses on the positive Boolean indicators are averaged over the five legal scenarios.} \label{tab:res:agg:subdim}
\begin{tabular}{lcccc}
\hline
\textbf{Model} & \textbf{D1} & \textbf{D2} & \textbf{D3} & \textbf{D4} \\
 & Role Adh. & Arg. Depth & Fact. Cons. & Ctx. Rel. \\
\hline
Llama-3.1-8B   & 0.93 & 0.97  & 0.80 & 1.00 \\
Gemma-2-9B-it  & 0.93 & 0.91  & 0.73 & 1.00 \\
Qwen-3-8B      & 0.90 & 0.94  & 0.73 & 1.00 \\
Gemma-3-12B-it & 0.73 & 1.00  & 0.73 & 0.83 \\
Hermes-3-8B    & 0.80 & 0.94  & 0.80 & 0.70 \\
Phi-4-14B      & 0.66 & 0.82  & 0.73 & 0.83 \\
Qwen-3-14B     & 0.46 & 1.00  & 0.63 & 0.83 \\
\hline
\end{tabular}
\end{table}

The analysis for each individual scenario highlights internal variability. Phi-4-14B records the lowest per-case score of the entire sample in Case 4 (0.28), with a complete reversal of procedural roles that compromises the overall consistency of the simulation, despite optimal performance in Cases 1 and 2. Similarly, Gemma-3-12B-it fluctuates between 0.54 in Case 4 and 1.00 in Cases 2 and 3, suggesting dependence on the specific narrative context. Llama-3.1-8B emerges for its greater inter-scenario regularity, with scores ranging from 0.84 to 1.00, albeit at the cost of three restarts distributed across different cases.
\begin{table*}
\centering
\caption{Behavioural Role Fidelity Quality Score ($QS_{d_{BRF}}$) and Retry Rate ($R$) in each individual legal scenario.} \label{tab:res:perscenario}
\begin{tabular}{lcccccccccc}
\hline
&
\multicolumn{2}{c}{\textbf{Case 1}} &
\multicolumn{2}{c}{\textbf{Case 2}} &
\multicolumn{2}{c}{\textbf{Case 3}} &
\multicolumn{2}{c}{\textbf{Case 4}} &
\multicolumn{2}{c}{\textbf{Case 5}} \\
 \textbf{Model} & $QS_{d_{BRF}}$ & $R$ & $QS_{d_{BRF}}$ & $R$ & $QS_{d_{BRF}}$ & $R$ & $QS_{d_{BRF}}$ & $R$ & $QS_{d_{BRF}}$ & $R$ \\
\hline
Llama-3.1-8B   & 0.96  & 0 & 0.96  & 0 & 0.88  & 1 & 0.84  & 1 & 1.00 & 1 \\
Gemma-2-9B-it  & 0.96  & 0 & 0.88  & 0 & 0.79  & 0 & 1.00 & 0 & 0.84  & 0 \\
Qwen-3-8B      & 1.00 & 0 & 0.88  & 0 & 0.88  & 1 & 0.92  & 0 & 0.79  & 0 \\
Gemma-3-12B-it & 0.79 & 0 & 1.00 & 0 & 1.00 & 0 & 0.54  & 0 & 0.79  & 0 \\
Hermes-3-8B    & 1.00 & 0 & 0.75  & 2 & 0.83  & 1 & 0.55  & 0 & 0.92  & 0 \\
Phi-4-14B      & 1.00 & 0 & 0.92  & 0 & 0.70  & 0 & 0.28  & 0 & 0.92  & 0 \\
Qwen-3-14B     & 0.87 & 0 & 0.87  & 0 & 0.50  & 0 & 0.71  & 0 & 0.71  & 0 \\
\hline
\end{tabular}
\end{table*}

To provide a complementary perspective on the evaluation dimensions assessed by the RPA-Check framework, we additionally report a decomposed analysis separating the Linguistic Formalism and Orthography Quality Score ($QS_{d_{LFO}}$) dimension from the Procedural Convergence and Stability Quality Score($QS_{d_{PCS}}$) dimension. The related results are summarized in Table~\ref{tab:res:agg2}. All evaluated models achieve strong performance on the $QS_{d_{LFO}}$ dimension, with four models reaching the maximum score (1.00) and the remaining three scoring 0.97, 0.97, and 0.90 respectively. In fact, as highlighted in Table~\ref{tab:res:perscenario2}, only QWEN-3-14B in Case 1, 2, and 3 and Hermes-3-8B in Case 3 and 4 score a $QS_{d_{LFO}}$ below 1.00, differently from all the other models across all the cases. This uniformly high linguistic fidelity confirms that quantized local models, regardless of parameter scale, are capable of maintaining grammatical precision and domain-appropriate register across forensic dialogue sessions. In evaluating LLM Court, therefore, the discriminative power of the framework is concentrated in the $QS_{d_{PCS}}$ dimension, where inter-model variance is substantially higher, ranging from 0.14 (Phi-4-14B) to 0.97 (Gemma-2-9B-it).

\begin{table}
\centering
\caption{Decomposed performance of models across the five legal scenarios, reporting scores for the Linguistic Formalism and Orthography Quality Score ($QS_{d_{LFO}}$) and Procedural Convergence and Stability Quality Score($QS_{d_{PCS}}$) dimensions independently.} \label{tab:res:agg2}
\begin{tabular}{lccc}
\hline
\textbf{Model} & \textbf{$QS_{d_{LFO}}$} & \textbf{$QS_{d_{PCS}}$} \\
\hline
Llama-3.1-8B    & 1.00 & 0.94 \\
Gemma-2-9B-it   & 1.00 & 0.97 \\
Qwen-3-8B       & 0.97  & 0.83 \\
Gemma-3-12B-it  & 1.00 & 0.66 \\
Hermes-3-8B     & 0.97  & 0.74 \\
Phi-4-14B       & 1.00 & 0.54 \\
Qwen-3-14B      & 0.90  & 0.86 \\
\hline
\end{tabular}
\end{table}

The per-scenario breakdown of $QS_{d_{PCS}}$, reported in Table~\ref{tab:res:perscenario3}, further reveals the sensitivity of procedural stability to specific narrative contexts. Phi-4-14B, for instance, records optimal performance in Cases 1 and 2 (1.00 in both) but degrades markedly in Cases 3, 4, and 5, reaching scores of 0.43, 0.14, and 0.14 respectively. Gemma-3-12B-it exhibits a comparable pattern of inter-scenario inconsistency, oscillating between 0.14 in Case 5 and 1.00 in Cases 2 and 3. Conversely, Gemma-2-9B-it demonstrates the most stable per-scenario profile within the $QS_{d_{PCS}}$ dimension, achieving 0.86 in Case 1 and 1.00 in Case 2, 3, 4, and 5, recording zero restarts, which collectively establish it as the most procedurally reliable architecture tested. The impact of generation seed on operational stability is directly observable in the Hermes-3-8B results: in Case 2, a seed change enabled the model to recover from complete failure to full procedural success, whereas in Case 3 the same intervention yielded only a very small improvement, suggesting that narrative complexity interacts non-linearly with model stochasticity.

\begin{table}
\centering
\caption{Linguistic Formalism and Orthography Quality Score ($QS_{d_{LFO}}$) for each individual legal scenario.} \label{tab:res:perscenario2}
\begin{tabular}{lcccccccccc}
\hline
& \textbf{Case 1} & \textbf{Case 2} & \textbf{Case 3} & \textbf{Case 4} & \textbf{Case 5} \\
\textbf{Model} & $QS_{d_{LFO}}$ & $QS_{d_{LFO}}$ & $QS_{d_{LFO}}$ & $QS_{d_{LFO}}$ & $QS_{d_{LFO}}$ \\
\hline
Llama-3.1-8B   & 1.00 & 1.00 & 1.00 & 1.00 & 1.00 \\
Gemma-2-9B-it  & 1.00 & 1.00 & 1.00 & 1.00 & 1.00 \\
Qwen-3-8B      & 1.00 & 1.00 & 1.00 & 1.00 & 0.93  \\
Gemma-3-12B-it & 1.00 & 1.00 & 1.00 & 1.00 & 1.00  \\
Hermes-3-8B    & 1.00 & 1.00 & 0.83  & 0.83  & 1.00  \\
Phi-4-14B      & 1.00 & 1.00 & 1.00 & 1.00 & 1.00  \\
Qwen-3-14B     & 0.83  & 0.83  & 0.83  & 1.00 & 1.00 \\
\hline
\end{tabular}
\end{table}

\begin{table}
\centering
\caption{Procedural Convergence and Stability Quality Score ($QS_{d_{PCS}}$) for each individual legal scenario.} \label{tab:res:perscenario3}
\begin{tabular}{lcccccccccc}
\hline
& \textbf{Case 1} & \textbf{Case 2} & \textbf{Case 3} & \textbf{Case 4} & \textbf{Case 5} \\
\textbf{Model} & $QS_{d_{PCS}}$ & $QS_{d_{PCS}}$ & $QS_{d_{PCS}}$ & $QS_{d_{PCS}}$ & $QS_{d_{PCS}}$ \\
\hline
Llama-3.1-8B   & 1.00 & 1.00 & 1.00 & 0.71  & 1.00 \\
Gemma-2-9B-it  & 0.86  & 1.00 & 1.00 & 1.00 & 1.00 \\
Qwen-3-8B      & 0.29  & 1.00 & 0.89  & 1.00 & 1.00  \\
Gemma-3-12B-it & 0.29  & 1.00 & 1.00 & 0.71  & 0.14  \\
Hermes-3-8B    & 1.00 & 1.00 & 0.14  & 1.00 & 0.43  \\
Phi-4-14B      & 1.00 & 1.00 & 0.43  & 0.14  & 0.14  \\
Qwen-3-14B     & 0.43  & 1.00 & 1.00 & 1.00 & 1.00 \\
\hline
\end{tabular}
\end{table}

\section{Discussion}\label{sec:disc}
Concerning the Behavioural Role Fidelity, the most relevant result is the emergence of an inverse relationship between the model's parametric size and procedural consistency. Contrary to expectations based on generalist benchmarks, such as MMLU~\citep{Hendrycks2021} or Chatbot Arena~\citep{Chiang2024}, smaller models (8–9 billion parameters), when properly instruction-tuned, demonstrate superior behavioral stability in environments with strong role constraints. Llama-3.1-8B, Gemma-2-9B-it, and Qwen-3-8B maintain high and consistent scores on D1 and D4, the sub-dimensions that measure functional adherence to the role and logical continuity of dialogue, respectively. This observation is consistent with literature on instruction tuning~\citep{Ouyang2022,Liu2024}, which shows that the quality of behavioral alignment depends more on the quality of the training data than on the scale of the model, and is particularly relevant in application contexts with limited computational resources.

The trade-off between Argumentative Depth and Role Adherence, quantified in Table~\ref{tab:res:agg:subdim}, is also relevant. The larger models, in particular Qwen-3-14B and Gemma-3-12B-it, generate complex argumentative responses, as evidenced by the maximum scores on D2, but produce frequent deviations from the assigned procedural constraints, with Role Adherence at 0.46 and 0.73, respectively. This behavior reveals the phenomenon of User-Alignment Bias (Sycophancy): models optimized to maximize user satisfaction tend to sacrifice adherence to the adversarial role of the Prosecutor or the institutional neutrality of the Judge in favor of dialogically rich but procedurally inconsistent responses.

Sensitivity to generation seeds and scenario specificity is another limitation that emerged from the per-case analysis and that the proposed framework is able to catch. In fact, the Retry Rate is not evenly distributed across models, but is concentrated on particular model-scenario combinations, suggesting that the variability inherent in autoregressive models interacts with the narrative complexity of the scenario in a non-linear way. Hermes-3-8B, for example, in addition to the 1.00 $QS_{d_{BRF}}$ in Case 1 that drops to 0.55 in Case 4, had three restart (two in Case 2 and 1 in Case 3) suggesting catastrophic events, such as the freezing of the game due to the incapability of tagging the next RPA in the conversation. This indicates that some case configurations trigger narrative degradation modes that are difficult to prevent through prompt optimization alone. This observation leads to an important methodological consideration about the proposed framework: the $R$ should not be interpreted as a simple indicator of model failure, but as a metric of stochastic instability which, when integrated with the $QS_{d_{BRF}}$, provides a more complete characterization of agent behavior in multi-turn scenarios.

The results in Table~\ref{tab:res:agg2} yield an additional finding of methodological relevance: the $QS_{d_{LFO}}$ dimension acts as a near-uniform ceiling across all evaluated architectures and is therefore insufficient as a standalone discriminator for model selection in constrained role-playing environments. The near-perfect $QS_{d_{LFO}}$ scores (0.90--1.00) observed across all seven models confirm that quantized local models are uniformly capable of sustaining the linguistic register required by forensic simulations.

The $QS_{d_{PCS}}$ dimension, by contrast, reveals a substantially differentiated performance landscape and constitutes the primary locus of discriminative power within the framework. Notably, the ranking produced by $QS_{d_{PCS}}$ diverges from that produced by the aggregate $QS_{d_{BRF}}$  in Table~\ref{tab:res:agg}: Gemma-2-9B-it achieves the highest dPCS score (0.97) while ranking second in aggregate $QS_{d_{BRF}}$ (0.89), whereas Llama-3.1-8B, which leads in aggregate $QS_{d_{BRF}}$ (0.925), records a $QS_{d_{PCS}}$ score (0.94) at the cost of three restarts. This divergence highlights the complementary nature of the $QS_{d_{BRF}}$, $QS_{d_{LFO}}$, $QS_{d_{PCS}}$  and $R$ metrics within the RPA-Check framework: a model may attain high narrative quality on completed interactions while simultaneously exhibiting elevated stochastic instability that necessitates manual intervention. By explicitly separating linguistic fidelity from procedural fidelity through distinct dimensions, the framework avoids the conflation of fluency with behavioral correctness that would arise from single-score evaluation paradigms. For deployment contexts where operational continuity is a hard constraint, such as autonomous educational platforms without human oversight, Gemma-2-9B-it's combination of zero restarts and near-ceiling $QS_{d_{PCS}}$ performance renders it the preferable candidate despite its marginally lower aggregate $QS_{d_{BRF}}$.

A particularly counterintuitive finding concerns the inverse relationship between model scale and $QS_{d_{PCS}}$ performance. Among the three largest models evaluated (Gemma-3-12B-it, Qwen-3-14B, and Phi-4-14B, all at or above 12 billion parameters), $QS_{d_{PCS}}$ scores cluster in the range 0.54--0.86, consistently below the 0.94--0.97 range recorded by Llama-3.1-8B and Gemma-2-9B-it. This pattern is consistent with the User-Alignment Bias (Sycophancy) effect identified for the Behavioral Role Fidelity dimension: models trained at larger scale with extensive instruction-following optimization appear to prioritize dialogic richness and cooperative responsiveness over strict procedural compliance, a tendency that is particularly detrimental in the verdict generation phase. In LLM Court, the verdict phase requires the model to produce a structurally constrained output, a binary outcome label combined with a coherent justification grounded in accumulated dialogue history, and it is precisely here that larger models exhibit the highest failure rates, either omitting explicit verdict labels or emitting verdicts semantically inconsistent with the preceding argumentation.

The per-scenario analysis in Table~\ref{tab:res:perscenario3} further substantiates the framework's capacity to expose model-specific failure modes that aggregate metrics would otherwise obscure. Phi-4-14B presents the starkest example: its $QS_{d_{PCS}}$ maximum scores (1.00) in Cases 1 and 2 collapse to 0.14 in Cases 4 and 5, a variance of 86 percentage points that cannot be attributed to a systematic architectural deficit but rather to sensitivity to narrative context. This high inter-scenario variance, detected and quantified through the RPA-Check checklist structure, underscores the necessity of multi-scenario evaluation corpora when benchmarking RPAs: single-scenario assessments would systematically overestimate or underestimate model reliability depending on whether the selected scenario happens to trigger narrative degradation modes in the evaluated architecture. The framework's granular Boolean checklist structure is thus not only a tool for measuring performance but also a diagnostic instrument capable of distinguishing between models with consistently moderate performance and models with high-variance profiles that may be unsuitable for deployment despite occasional peak results.

The scenario-level results further suggest that narrative complexity and evidentiary ambiguity, as defined in the legal scenario corpus, are meaningful predictors of model-specific failure modes. Case 4 characterized by contested boundary evidence and witness bias in a non-criminal civil proceeding, consistently elicits the lowest $QS_{d_{BRF}}$ scores across multiple architectures, Phi-4-14B (0.28), Gemma-3-12B-it (0.54), and Hermes-3-8B (0.55), suggesting that the adversarial civil context may systematically destabilize role adherence in ways that criminal scenarios do not. Conversely, Case 5, the most demanding scenario in terms of argumentative depth and long-term narrative stability due to its multiplicity of witnesses and inherently ambiguous testimony, does not uniformly produce the lowest scores, indicating that parametric scale and instruction tuning interact differently with evidentiary complexity than with procedural novelty. These observations reinforce the necessity of scenario corpus diversity in RPA benchmarking, as scenario-specific structural properties, rather than overall difficulty, appear to be the primary triggers of architecture-dependent degradation.

\subsection{Limitations}\label{subsec:disc:limit}

The main limitation of the proposed methodology is in its experimental evaluation: a corpus of five scenarios per model constitutes a relatively limited sample, which constrains the statistical significance of inter-scenario variance estimates. The per-case results in Table~\ref{tab:res:perscenario} do reveal meaningful behavioral differences across narrative contexts, and the scenario-level analysis remains informative for identifying model-specific failure modes; however, broader generalization of the quantitative findings would benefit from a larger, automatically generated case corpus. This is a natural direction for future work, given that the procedural case generation module described in Section~\ref{sec:case_gen} can be extended to produce such a corpus in a reproducible manner.

This limitation defines the conditions under which the experimental results should be interpreted. The framework demonstrates its intended capability, i.e., transforming qualitative behavioral requirements into verifiable quantitative metrics and supporting reproducible comparative analysis across heterogeneous architectures, within these boundaries. Specifically, the results show that the framework can discriminate between distinct performance profiles at a diagnostic level, distinguishing, for instance, a model that maintains high sequential consistency (D4) from one that prioritizes argumentative density (D2). This granularity is relevant for model selection in constrained simulation environments, even as the absolute scores should be treated as indicative rather than definitive benchmarks pending evaluation on a larger scenario corpus.

\section{Conclusions}\label{sec:conc}
This paper presented RPA-Check, a multi-stage automated evaluation framework that addresses a critical methodological gap in the assessment of LLM-based Role-Playing Agents operating under strict procedural and semantic constraints. The framework transforms qualitative behavioral requirements into granular, verifiable Boolean metrics, enabling reproducible and objective comparison across heterogeneous model architectures. Validation through LLM Court, a serious game that is a forensic simulation environment confirmed the framework's discriminative capacity across three primary evaluation dimensions: Behavioral Role Fidelity, Procedural Convergence and Stability, and Linguistic Formalism and Orthography.

Specifically, the experimental evidence collected from LLM court using RPA-Check establishes a principled and empirically grounded counterpoint to the prevailing assumption that parametric scale is a sufficient proxy for agent capability in constrained generative environments. The inverse relationship between model scale and procedural consistency, most sharply manifested in the sycophancy-driven role deviations of the larger architectures, underscores that instruction tuning quality and behavioral alignment methodology are more determinative of operational reliability than parameter count alone. Equally significant is the near-uniform ceiling effect observed across all evaluated models on the linguistic formalism dimension, which confirms that discriminative evaluation power in specialized role-playing contexts resides not in surface fluency but in the capacity to sustain procedural and semantic coherence across extended, adversarially structured interactions. Moreover, the high inter-scenario variance detected in architectures (such as Phi-4-14B and Gemma-3-12B-it) indicates that single-scenario benchmarking constitutes a structurally inadequate evaluation protocol for RPAs, motivating the construction of larger, automatically generated case corpora using the procedural generation pipeline introduced here.

More broadly, the modular architecture of RPA-Check (where dimensions, augmentation strategies, and filtering criteria are domain-configurable) makes the framework a transferable tool for evaluating constrained generative agents across specialized domains, potentially beyond the forensic context (such as in medical simulation, compliance training, and regulated negotiation environments).



\appendix

\section{RPA-Check Evaluation Prompts}\label{appendix:prompts}

The complete set of prompts employed in the RPA-Check evaluation pipeline as applied to the LLM Court environment is reported hereby. The prompts implement \textit{Stage 2}, \textit{Stage 3}, and \textit{Stage 4} of the proposed framework: Augmentation (via Question Diversification and Question Elaboration), Semantic Filtering, and LLM-as-a-Judge Evaluation. Each prompt is designed to transform qualitative behavioral dimensions into granular Boolean indicators and to ensure objectivity, non-redundancy, and agent isolation across the checklist generation and scoring stages. 

Specifically, the Augmentation – Question Diversification prompt instructs the generator model ($\mathcal{M}_{gen}$,  equation \ref{eq:generator}) to expand seed questions into alternative formulations that explore the same evaluation criterion from multiple perspectives, ensuring broader conceptual coverage. The Augmentation – Question Elaboration prompt decomposes each seed question into a hierarchy of five to ten sub-questions, increasing the resolution of the evaluation at the level of individual sub-dimensions.

The Filtering prompt applies the three retention criteria  (i.e., dimension alignment, non-redundancy, and stylistic adequacy) to the raw question pool, removing items that deviate from the target dimension definition, overlap semantically with retained questions, or introduce exaggerated or overly narrow formulations.

Finally, the Evaluation prompt presents the filtered checklist to the judge model ($\mathcal{M}_{judge}$, equation \ref{eq:judge}) alongside the case description and conversation history, requiring binary yes/no responses with inline Chain-of-Thought justifications while explicitly excluding player-controlled dialogue turns from the scoring process.

\begin{lstlisting}[label=QuestDiv,caption=AUGMENTATION – QUESTION DIVERSIFICATION PROMPT]
<Task Overview>
You will be provided with: 1) Information about the benchmark to be evaluated,2) The main concept being assessed in the benchmark, and 3) Seed questions that include key components and sub
questions related to this concept.
Your task is to create additional sub-questions for the key components to comprehensively assess the main concept. Each sub-question must meet givenconditions to ensure a high-quality question set.
 
 1) Benchmark Information:
 LLM Court: a benchmark that evaluates dialogues on different characteristics to extract the level of ability of impersonation of models in legal contexts.
 
 2) Main Concept in the Benchmark:
 {concept: role-play}: {description: the model's ability to impersonate characters in a legal context}
 
 3) Key Components and Seed Questions:
{seed questions}
 
<Conditions for a Good Question List>
- Concepts included in different questions should not be like each other
- Please provide me with the diversification by translating the respective questions into English.

<Constraints>
- Each sub-question must be answerable with a simple "yes" or "no".
- A "yes" answer should indicate that the sentence improves the specified evaluation criterion (e.g., Coherence, Relevance).
- Each question should assess only a single dimension or concept.
- Each question should not ask about more than one topic or concept.
\end{lstlisting}

\begin{lstlisting}[label=QuestAug,caption=AUGMENTATION – QUESTION ELABORATION PROMPT]
<Task Overview>
Your task is to generate multiple additional questions to evaluate benchmark performance under specific constraints. 
You will receive the key component and sub-component evaluating {Behavioral Role Fidelity} and the question related to it. 
The definition of {Behavioral Role Fidelity} is as follows: {This dimension evaluates the content of the characters' answers both in their ability to follow their role and keep the answers deep and interesting.}
The evaluation for dimension {Behavioral Role Fidelity} will be centered around the key component {Role Adherence, Argumentative Depth, Factual Consistency, Contextual Relevance}.

 <TASK>
 # Your role: You have to break down sub-questions into 3 to 10 sub-sub-questions considering {Behavioral Role Fidelity} when pairs of seed name and question are given.
 # Benchmark information: The benchmark consists of courtroom-style dialogues generated by LLMs. The goal is to evaluate whether the response maintain character role fidelity, provide substantial and relevant content, avoid internal contradictions, and follow a logical conversational flow.

 <CONSTRAINTS>
- Each sub-question must be answerable with a simple "yes" or "no".
- A "yes" answer should indicate that the sentence improves the specified evaluation criterion.
- Each question should assess only a single dimension or concept.
- Each question should not ask about more than one topic or concept.

 <Conditions for a Good Question List>
- Questions must be clear, specific, and unambiguous.
- Avoid vague or overly broad formulations.
- Ensure that questions are grounded in the definition of the dimension.
- Avoid redundancy and overlapping meaning between questions.
- The role of the defense is represented by the player and not by the LLMs
 
 <FORMAT>
1. Answers' coherence with roles:
1-1. Are the lines consistent with the corresponding role?
1-1-1. q1-1-1_aug_question
 1-1-2. q1-1-2_aug_question
 
2. Content depth:
1-1. Do the lines have substantial content?
1-1-1. q1-1-1_aug_question
 1-1-2. q1-1-2_aug_question
 
3. Contradictions:
1-1. Are there contradictions in the lines?
1-1-1. q1-1-1_aug_question
 1-1-2. q1-1-2_aug_question
 
4. Logical connection between lines:
1-1. Are the lines logically connected to the previous ones?
1-1-1. q1-1-1_aug_question
 1-1-2. q1-1-2_aug_question
 
<EXAMPLE>
{example}

\end{lstlisting}

\begin{lstlisting}[label=Filtering,caption=FILTERING PROMPT]
<Task Overview>
 Your task is to filter out questions from a list based on the following criteria:
 
 1) dimension Alignment:
- dimension definition: {Behavioral Role Fidelity: This dimension evaluates the content of the characters' answers both in their ability to follow their role and keep the answers deep and interesting}
- Remove questions that deviate from the given dimension's definition.
- Remove questions that are more closely related to other dimensions than the current one.
 
 2) Redundancy:
- Remove questions that:
 * Ask for the same or very similar information (even if phrased differently).
 * Convey very similar meanings without adding unique insight.
 
 3) Style:
- Remove questions that:
 * Use overly exaggerated wording.
 * Focus on excessively detailed or minor points that don't meaningfully affect overall quality.
 
 4) Benchmark Context
- Name: LLM Court 
- Purpose: Evaluation of role-playing capabilities 
- Key Metrics: Naturalness, Coherence, Engagingness, Groundedness
- Do not modify any of the remaining questions or generate new ones.
- Keep questions in their original dictionary format.
 
 5) Sub-dimensions and Questions:
{format_sub_dimensions(sub_dimensions)}
 
6) Output Requirements:
- Output format: JSON only
- Structure: 
{" Answers coherence with roles": [
 "Filtered Question 1",
 "Filtered Question 2"]}
  
{" Content depth ": [
 "Filtered Question 1",
 "Filtered Question 2"]}
 
{" Contradictions": [
 "Filtered Question 1",
 "Filtered Question 2"]}

{" Logical connection between lines ": [
 "Filtered Question 1",
 "Filtered Question 2"]}
 
{" Removed questions": [
 "Removed Question 1",
 "Removed Question 2"]}

 <Important Note>
- Do not modify the content of remaining questions
- Do not generate new questions
- Maintain the original dictionary format
- Only remove questions that fail the above criteria
- Do not remove entire sub-dimensions or their keys unless no valid questions remain

\end{lstlisting}

\begin{lstlisting}[label=Eval,caption=EVALUATION PROMPT]
<Task Overview>
You will be given a conversation between different characters of a courtroom. You will then be given
the courtroom case description and the dialogue spoken by the characters of the courtroom. Your task is to read the provided conversation history and the courtroom description, then answer "yes" or "no" to specific questions. These questions will relate to a particular dimension of the conversation.
 
<dimension Definition>
This dimension evaluates whether each character consistently behaves according to their assigned role (e.g., witnesses provide testimony, prosecutor argues against the accused, judge remains neutral and procedural). Any deviations from the expected behavior (such as a witness asking questions like a prosecutor) must be flagged and considered a break in role coherence, regardless of content quality.
 
<Instructions>
1. Read these instructions thoroughly.
2. Carefully read the Case Description and the Conversation History.
3. Understand the given questions and the definition of the Behavioral Role Fidelity.
4. Respond to each question with "yes" or "no". Base your answers on a clear rationale.
5. Follow the specified format for your answers.
6.The defense is controlled by the player. Do not include any dialogue from this role in your evaluation, regardless of how it is labeled (e.g., "Defense", "Defense Attorney", "Player's Lawyer", or any other variant).
If a turn's content clearly indicates it is the player-controlled defense (e.g., giving closing arguments, objecting, questioning witnesses), also exclude it, even if the tag is missing or ambiguous.
Only evaluate lines spoken by AI-controlled characters (Prosecutor, Judge, Witnesses, etc.).
7. The system message is a description of what the character needs to do, you must avoid including them in your answering because the benchmark is evaluating the AI.
8. Extra emphasis: Before evaluating content depth or logical connections, first check that each character stays in their assigned role. If a character speaks or acts outside their role (e.g., a witness interrogates another witness, a judge gives opinions on guilt), mark this as a coherence issue and note it in your reasoning.
 
 <Answer Format>
Q1 - [Question]: [Your Answer]
Q2 - [Question]: [Your Answer]

Include a brief justification in parentheses after each yes/no, e.g., 'Q1 - [Question]: [Your Answer] (reasoning: ...)'.
 ...
 # Case Description #
<case description>
 
 # Conversation History #
 <dialogue>
 
 # Questions #
 <questions>

 # Your Answer #
 Provide your answers to the given questions, following the specified Answer Format.

\end{lstlisting}

\printcredits

\bibliographystyle{cas-model2-names}

\bibliography{mybib}



\end{document}